\def\year{2022}\relax
\documentclass[letterpaper]{article} 
\usepackage{aaai22}  
\usepackage{times}  
\usepackage{helvet}  
\usepackage{courier}  
\usepackage[hyphens]{url}  
\usepackage{graphicx} 
\usepackage{natbib}  
\usepackage{caption} 
\DeclareCaptionStyle{ruled}{labelfont=normalfont,labelsep=colon,strut=off} 
\usepackage{algorithm}
\usepackage{algorithmic}

\usepackage{newfloat}
\usepackage{listings}
\floatstyle{ruled}
\newfloat{listing}{tb}{lst}{}
\floatname{listing}{Listing}
\title{KOALA: A Kalman Optimization Algorithm with Loss Adaptivity}
\author {
    Aram Davtyan, Sepehr Sameni, Llukman Cerkezi, Givi Meishvilli, Adam Bielski, Paolo Favaro
}
\affiliations {
    Computer Vision Group, University of Bern \\
    \{firstname.lastname\}@inf.unibe.ch
}

\usepackage{times}
\usepackage{epsfig}
\usepackage{graphicx}
\usepackage{amsmath}
\usepackage{amssymb}
\usepackage{booktabs}
\usepackage{multirow}
\usepackage{algorithm}
\usepackage{algorithmic}
\usepackage{subfig}
\usepackage{nicefrac}
\usepackage{xcolor}
\usepackage{rotating}

\usepackage{amsthm}
\newtheorem{theorem}{Theorem}

\newcommand{\ie}{\emph{i.e.}}
\newcommand{\eg}{\emph{e.g.}}
\newcommand{\etal}{{et al.}}

\begin{document}

\maketitle

\begin{abstract}
   Optimization is often cast as a deterministic problem, where the solution is found through some iterative procedure such as gradient descent. However, when training neural networks the loss function changes over (iteration) time due to the randomized selection of a subset of the samples. This randomization turns the optimization problem into a stochastic one.
   We propose to consider the loss as a noisy observation with respect to some reference optimum. This interpretation of the loss allows us to adopt Kalman filtering as an optimizer, as its recursive formulation is designed to estimate unknown parameters from noisy measurements. 
   Moreover, we show that the Kalman Filter dynamical model for the evolution of the unknown parameters can be used to capture the gradient dynamics of advanced methods such as Momentum and Adam. 
   We call this stochastic optimization method KOALA, which is short for Kalman Optimization Algorithm with Loss Adaptivity. KOALA is an easy to implement, scalable, and efficient method to train neural networks. We provide convergence analysis and show experimentally that it yields parameter estimates that are on par with or better than existing state of the art optimization algorithms across several neural network architectures and machine learning tasks, such as computer vision and language modeling.
\end{abstract}

\section{Introduction}

Optimization of functions involving large datasets and high dimensional models finds today large applicability in several data-driven fields in science and the industry. 
Given the growing role of deep learning, in this paper we look at optimization problems arising in the training of neural networks. The training of these models can be cast as the minimization or maximization of a certain objective function with respect to the model parameters. Because of the complexity and computational requirements of the objective function, the data and the models, 
the common practice is to resort to iterative training procedures, such as gradient descent.
Among the iterative methods that emerged as the most effective and computationally efficient is stochastic gradient descent (SGD) \cite{SGD}. SGD owes its performance gains to the adoption of an approximate version of the objective function at each iteration step, which, in turn, yields an approximate or noisy gradient.

While SGD seems to benefit greatly (\eg, in terms of rate of convergence) from such an approximation, it has also been shown that too much noise hurts the performance \cite{wang2013variance,bottou2018optimization}.
This suggests that, to further improve over SGD, one could attempt to model the noise of the objective function. 
We consider the iteration-time varying loss function used in SGD as a stochastic process obtained by adding the empirical risk to zero mean Gaussian noise. A powerful approach designed to handle estimation with such processes is Kalman filtering~\cite{kalman1960new}. In fact, Kalman filtering has been used to train neural networks before \cite{haykin2004kalman, vivak2016,ISMAIL2018509}. However, it can be applied in very different ways. Indeed, in our approach, which we call \emph{KOALA}, we introduce a number of novel ideas that result in a practical and effective training algorithm. Firstly, we introduce drastic approximations of the estimated covariance of Kalman's dynamical state so that the corresponding matrix depends on only up to a $2\times 2$ matrix of parameters. Secondly, we approximate intermediate Kalman filtering calculations so that more accuracy can be achieved. Thirdly, because of the way we model the objective function, we can also define a schedule for the optimization that behaves similarly to learning rate schedules used in SGD and other iterative methods \cite{ICLR_KingmaB_2014}.\\
\textbf{Our contributions can be summarized as follows:} 1) We design KOALA so that it can 
handle high-dimensional data and models, and large datasets; 2) We present analysis and conditions to ensure convergence;
3) We allow both the automated tuning of the algorithm and also the use of a learning rate schedule similar to those in existing methods;
4) We incorporate the automatic adaptation to the noise in the loss, which might vary depending on the settings of the training (\eg, the minibatch size), and to the variation in the estimated weights over iteration time;
5) We show how to incorporate iteration-time dynamics of the model parameters, which are analogous to momentum \cite{pmlr_sutskever_2013};
6) We introduce KOALA as a framework so that it can be further extended (we show two variations of KOALA); 
7) We show experimentally that KOALA is on par with state of the art optimizers and can yield better minima at \textbf{test time} in a number of problems from image classification to generative adversarial networks (GAN) and natural language processing (NLP).

\section{Prior Work}


\noindent\textbf{First-Order Methods.} 
First-order methods exploit only the gradient of the objective function. 
The main advantage of these methods lies in their speed and simplicity. 
\cite{SGD} introduce the very first stochastic optimization method (SGD) in early 1951. 
Since then, the SGD method has been thoroughly analyzed and extended
\cite{shang2018vrsgd,NEURIPS2020_1cdf14d1,sung2020ssgd}. 
However, a limitation of SGD is that the learning rate must be manually defined and it does not take any measures to improve the gradient direction.\\ 
\noindent\textbf{Second-Order Methods.} 
To address the manual tuning of the learning rates in first-order methods and to improve the convergence rate, second-order methods rely on the Hessian matrix.
However, this matrix grows quadratically with the number of model parameters.
Thus, most work reduces the computational complexity by approximating the Hessian~\cite{NEURIPS2020_Goldfarb,BotevRB17_ICML}. 
A number of methods looks at combining the second-order information in different ways. 
For example, \cite{ICML_RouxF10_2010} combine Newton's method and natural gradient. \cite{pmlr_sohl_dicksteinb_2014} combine SGD with the second-order curvature information leveraged by quasi-Newton methods. 
\cite{yao2020adahessian} dynamically incorporate the curvature of the loss via adaptive estimates of the Hessian. 
\cite{Henriques_2019_ICCV} propose a method that does not require to store the Hessian at all. 
In contrast to these methods, KOALA does not compute second-order derivatives, but focuses on modeling noise in the objective function.\\ 
\noindent\textbf{Adaptive.} 
An alternative to using second-order derivatives is to automatically adjust the step-size during the optimization.
The adaptive selection of the update step-size has been based on several principles, including: the local sharpness of the loss function \cite{yue2021salr}, incorporating a line search approach \cite{NEURIPS2019_2557911c,NEURIPS2020_Mutschler,NIPS2015_Mahsereci}, the gradient change speed \cite{DUBEY},
a ``belief'' in the current gradient direction \cite{NEURIPS2020_Zhuang}, 
the linearization of the loss \cite{NEURIPS2018_Rolinek},
the per-component unweighted mean of all historical gradients \cite{daley2021expectigrad},
handling noise by preconditioning based on a covariance matrix \cite{ijcai2017_Yasutoshi}, 
learning the update-step size \cite{wu2020wngrad},
looking ahead at the sequence of fast weights generated by another optimizer \cite{NEURIPS2019_Zhang}. 
A new family of sub-gradient methods called AdaGrad is presented in \cite{JMLR_v12_duchi11a}. 
AdaGrad dynamically incorporates knowledge of the geometry of the data observed in earlier iterations. \cite{tieleman2012lecture} introduce RmsProp, further extended in \cite{mukkamala17a} with logarithmic regret bounds for strongly convex functions. 
\cite{adadelta} propose a per-dimension learning rate method for gradient descent called AdaDelta. 
\cite{ICLR_KingmaB_2014} introduce Adam, based on adaptive estimates of lower-order moments. 
A wide range of variations and extensions of the original Adam optimizer has also been proposed
\cite{ADAM_BS,ADAM_NC,ADAMP_SGDP,ADAM_W,chen2018on,Liu2020On,luo2018adaptive,Wang2020SAdam}. 
Recent work proposes to decouple the weight decay \cite{granziol2021sgd,ginsburg2020stochastic}. 
\cite{Chen_ijcai2020} introduces a partially adaptive momentum estimation method. 
Some recent work also focuses on the role of gradient clipping \etal~\cite{NEURIPS2020_Zhang,NEURIPS2020_b05b57f6}. 
In most prior work, adaptivity comes from the introduction of extra hyper-parameters. In our case, this property is a direct byproduct of the Kalman filtering framework.\\
\noindent\textbf{Kalman filtering.} The use of Kalman filtering theory and methods for the training of neural networks is not new. 
For example, \cite{ISMAIL2018509} relates to our KOALA-V as the authors also work with scalar measurements. However, our approach differs in several ways as we introduce a way to incorporate Momentum, learning rate scheduling, noise adaptivity and provide a convergence analysis.
More recently, 
\cite{shashua2019trust} incorporated Kalman filtering for Value Approximation in Reinforcement Learning. 
\cite{ollivier2019extended} recovered the exact extended Kalman filter equations from first principles in statistical learning: the Extended Kalman filter is equal to Amari's online natural gradient, applied in the space of trajectories of the system. 
\cite{devilmarest2020stochastic} applied the Extended Kalman filter to linear and logistic regressions. 
\cite{kalman_takenga} compared GD to methods based on either Kalman filtering  or the decoupled Kalman filter. 
To summarize, all of these prior Kalman filtering approaches either focus on a specific non-general formulation or face difficulties when scaling to high-dimensional parameter spaces of large-scale neural models.

\section{Risk Minimization through Kalman Filtering}


In machine learning, we are interested in minimizing the \emph{expected risk}
\begin{align}
    \min_{x\in R^n} L(x),\quad \text{where } L(x) \doteq E_{\xi\sim p(\xi)}[\ell(\xi;x)], \label{eq:expectedrisk}
\end{align}
with respect to some loss $\ell$ that is a function of both the data $\xi\in\mathbb{R}^d$ with $d$ the data dimensionality, $p(\xi)$ is the probability density function of $\xi$, and the model parameters $x \in \mathbb{R}^n$ (\eg, the weights of a neural network), where $n$ is the number of parameters in the model. We consider the big data case, which is of common interest today, where both $d\gg 1$ and $n\gg 1$ (\eg, in the order of $10^6$). For notational simplicity, we do not distinguish the supervised and unsupervised learning cases by concatenating all data into a single vector $\xi$ (\eg, in the case of image classification we stack in $\xi$ both the input image and the output label). 
In practice, we have access to only a finite set of samples and thus settle for the \emph{empirical risk} optimization
\begin{align}
    \min_{x\in R^n} \hat L(x), \text{where } \hat L(x) \doteq \frac{1}{m}\sum_{i=1}^{m} \ell(\xi_i;x),
    \label{eq:empiricalrisk}
\end{align}
and $\xi_i\sim p(\xi)$, for $i=1,\dots,m$, are our training dataset samples. 
The above risk is often optimized iteratively via a gradient descent method, because a closed form solution (\eg, as with least squares) is typically not available.

Moreover, since in current datasets $m$, the number of training samples, can be very large, the computation of the gradient of the empirical risk at each iteration is too demanding. To address this issue, the stochastic gradient descent (SGD) method \cite{SGD} minimizes the following \emph{minibatch risk} at each iteration time $k$
\begin{align}
    \hat L_{k}(x) \doteq \frac{1}{|{\cal C}_k|}\sum_{i\in {\cal C}_k}\ell(\xi_i;x), 
\end{align}
where ${\cal C}_k\subset [1,\dots,m]$ is a random subset of the dataset indices.
Given a random initialization for $x_0$, SGD builds a sequence $\{x_k\}_{k=0,\dots,T}$ by recursively updating the parameters $x_k$ so that they decrease the $k$-th loss $\hat L_{k}$, \ie, for $k=0,\dots,T-1$
\begin{align}
    x_{k+1} = x_{k} - \eta \nabla \hat L_{k}(x_{k}), \label{eq:sgd}
\end{align}
where $\nabla \hat L_{k}(x_{k})$ denotes the gradient of $\hat L_{k}$ with respect to $x$ and computed at $x_{k}$, and $\eta>0$ is the \emph{learning rate}, which regulates the speed of convergence.

\subsection{Modeling Noise in Risk Minimization}

In KOALA, we directly model the statistical properties of the minibatch risk $\hat L_{k}$ as a function of the empirical risk $\hat L$. To relate $\hat L_{k}$ to $\hat L$ we start by looking at the relation between $\hat L$ and the expected risk $L$. First, we point out that $\hat L$ is the sample mean of $L$. Then, we recall that, because of the central limit theorem, $\hat L$ converges to a Gaussian distribution with mean $L$ as $m\rightarrow \infty$. The same analysis can be applied to the minibatch risk $\hat L_{k}$. $\hat L_{k}$ is a sample mean of $\hat L$ and as $|{\cal C}_{k}|\rightarrow m$, $\hat L_{k}$ converges to $\hat L$. Thus, the distribution of each $\hat L_{k}$ will tend towards a Gaussian random variable with mean $\hat L$. Finally, we can write $\forall x$
\begin{align}
\hat L_{k}(x) \simeq \hat L(x)-v_k,    
\label{eq:gaussianrelation}
\end{align} 
where the scalar noise variable $v_k\sim {\cal N}(0,R_k)$, is a zero-mean Gaussian with variance $R_k$. Later, we will show how to obtain an online estimate of $R_k$. 

\subsection{Risk Minimization as Loss Adaptivity}

Consider a model with parameters $\hat x$. For example, $\hat x$ could be chosen as one of the solutions of the optimization \eqref{eq:empiricalrisk}, \ie, such that $\hat L(\hat x) = \min_x \hat L(x)$. However, more in general, one can define $\hat L(\hat x) \doteq \hat L^\text{target} $, for some feasible $\hat L^\text{target}$. Let us now define the problem of finding $x_k$ such that 
\begin{align}
\hat L_{k}(x_k) = \hat L^\text{target} - v_k,    
\label{eq:measurement}
\end{align}
for all $k$ and where $v_k$ depends on $\hat x$.
The above formulation allows us to also solve the optimization in \eqref{eq:empiricalrisk}. Rather than explicitly finding the minimum of a function, in KOALA we look for the model parameters that \emph{adapt} the minibatch risk to a given value on average. 
However, to solve \eqref{eq:empiricalrisk} we need $\min_x\hat L(x)$, which is unknown. As an alternative, we iteratively approximate $\min_x \hat L(x)$ with a sequence of $\hat L^\text{target}_{k}$ that converges to it.
For example, by applying Theorem~\ref{thm:convergence1} (see next sections), the approximation $\min_x \hat L(x) \simeq \hat L^\text{target}_{k} \doteq \hat L_{k}(x_k) - k^{-1}$ will ensure the convergence of KOALA as $k$ grows.


\subsection{Kalman Filtering for Stochastic Optimization}

Eq.~\eqref{eq:measurement} can be interpreted as a noisy observation of some unknown model parameters $x$, which we want to identify. Kalman filtering is a natural solution to this task. As discussed in the Prior Work section, there is an extensive literature on the application of Kalman filtering as a stochastic gradient descent algorithm. However, these methods differ from our approach in several ways. For instance, Vuckovic~\cite{vuckovic2018kalman} uses the gradients as measurements. Thus, this method requires large matrix inversions, which are not scalable to the settings we consider in this paper and that are commonly used in deep learning (see section 3.3 in \cite{vuckovic2018kalman}). KOALA works instead directly with the scalar risks $\hat L_{k}$ and introduces a number of computational approximations that make the training with large datasets and high dimensional data feasible. 

Let us model the uncertainty of the identified parameters $x_k$ as a Gaussian random variable with the desired target $\hat x_k$ as mean. 
Then, a dynamical system for a sequence $x_k$ is 
\begin{align}
    x_k & = x_{k - 1} + w_{k-1} \label{eq:vanilla_states}\\
    \hat L^\text{target}_{k} & = \hat L_{k}(x_k) + v_k. \label{eq:vanilla_measurements}
\end{align}
Here, $w_k\sim {\cal N}(0,Q_k)$ is modeled as a zero-mean Gaussian variable with covariance $Q_k$. The dynamical model implies that the mean of the state $x_k$ does not change when it has adapted the mean minibatch risk to the target observation $\hat L^\text{target}_{k}$.
The equations~\eqref{eq:vanilla_states} and \eqref{eq:vanilla_measurements} describe a dynamical system suitable for 
Kalman filtering \cite{kalman1960new}. 
For completeness, we briefly recall here the general equations for an Extended Kalman filter
\begin{align}
    x_k & = f_k(x_{k - 1}) + w_{k-1} \\
    z_k & = h_k(x_k) + v_k,
\end{align}
where $x_k$ are also called the \emph{hidden state}, $z_k \in \mathbb{R}^s$ are the observations, $f_k$ and $h_k$ are  functions that describe the state transition and the measurement dynamics respectively. The zero-mean Gaussian noises added to each equation must also be independent of the hidden state. 
The Extended Kalman filter infers optimal estimates of the state variables from the previous estimates of $x_k$ and the last observation $z_k$. Moreover, it also estimates the a posteriori covariance matrix $P_k$ of the state. This is done in two steps: Predict and Update, which we recall in  Table~\ref{t:kalmanfilter}.

\begin{table}[t!]
	\begin{center}
	\caption{Extended Kalman filter recursive equations for a posteriori state and covariance estimation.} \label{t:kalmanfilter}
	\begin{tabular}{l l} \toprule
		 \multirow{2}{*}{Predict:} & $\hat x_k = f_k(x_{k - 1})$\\
		 & $\hat P_k = F_k P_{k - 1} F_k^\top + Q_k$ \\
		 \midrule
		 \multirow{4}{*}{Update:} & $S_k = H_k \hat P_k H_k^\top + R_k$ \\
		 & $K_k = \hat P_k H_k^\top S_k^{-1}$ \\
		 & $x_k = \hat x_k + K_k (z_k - h_k(\hat x_k))$ \\
		 & $P_k = (I - K_k H_k) \hat P_k$ \\
		 \midrule
		 \multirow{2}{*}{with:} & $H_k \doteq \nabla h_k(\hat x_k)$\\
		 & $F_k \doteq \nabla f_k(x_{k-1})$\\
		 \bottomrule
	\end{tabular}
	\end{center}
\end{table}


If we directly apply the equations in Table~\ref{t:kalmanfilter} to our equations~\eqref{eq:vanilla_states} and \eqref{eq:vanilla_measurements}, we would immediately find that the posterior covariance $P_k$ is an $n\times n$ matrix, which would be too large to store and update for $n$ values used in practice. Hence, we approximate $P_k$ as a scaled identity matrix. 
Since the update equation for the posterior covariance requires the computation of $K_k H_k = P_k H_k^\top H_k S_k^{-1}$, we need to approximate $H_k^\top H_k$ also with a scaled identity matrix. 
We do this by using its largest eigenvalue, \ie,
\begin{align}
    H_k^\top H_k \approx \left| H_k \right| ^2 I_{n\times n} = \left|\nabla_x \hat L_{k}(\hat x_k) \right| ^2 I_{n\times n},
\end{align}
where $I_{n\times n}$ denotes the $n\times n$ identity matrix.
Because we work with a scalar loss $\hat L_{k}$, the innovation covariance $S_k$ is a scalar and thus it can be easily inverted. 
The general framework introduced so far 
is very flexible and allows several extensions. 
The parameter estimation method obtained from equations~\eqref{eq:vanilla_states} and \eqref{eq:vanilla_measurements} is a special case of KOALA that we call \emph{KOALA-V} (Vanilla), and summarize in Algorithm~\ref{alg:vanillakalman}.

\begin{algorithm}[t]
    \caption{KOALA-V (Vanilla)}\label{alg:vanillakalman}
    \begin{algorithmic}
        \STATE Initialize $x_0$, $P_0$, $Q$ and $R$
        \FOR{$k$ in range(1, $T$)}
        \STATE Predict:
        \STATE $\quad \hat x_k = x_{k - 1}; \; \hat P_k = P_{k - 1} + Q$
        \STATE Update:
        \begin{align}
        x_k & = \hat x_k - \textstyle \frac{\hat P_k (\hat L_{k}(\hat x_k)-\hat L^\text{target}_{k})}{\hat P_k |\nabla \hat L_{k}(\hat x_k)|^2 + R} \nabla \hat L_{k}(\hat x_k) \label{eq:vanillakalman_update} \\
        P_k & = \textstyle \frac{ R}{\hat P_k |\nabla \hat L_{k}(\hat x_k)|^2 + R}\hat P_k
        \end{align}
        \ENDFOR
    \RETURN $x_K$
    \end{algorithmic}
\end{algorithm}

Notice that the update in eq.~\eqref{eq:vanillakalman_update} is quite similar to the SGD update~\eqref{eq:sgd}, where the learning rate $\eta$ depends on $\hat P_k$, $\hat L_{k}(\hat x_k)-\hat L^\text{target}_{k}$, $\nabla \hat L_{k}(\hat x_k)$ and $R$. Thus, the learning rate in KOALA-V automatically adapts over time to the current loss value, its gradient and estimation of the parameters, while in SGD it must be manually tuned.

\subsection{Incorporating Momentum Dynamics}

A first important change we introduce is the incorporation of Momentum \cite{pmlr_sutskever_2013}. Within our notation, this method could be written as
\begin{align}
    p_k & = \kappa p_{k - 1} - \eta \nabla \hat L(x_{k - 1}) \\
    x_k & = x_{k - 1} + p_{k},
\end{align}
where $p_k$ are so called \emph{momentums} or velocities, that accumulate the gradients from the past. The parameter $\kappa\in(0,1)$, commonly referred to as \emph{momentum rate}, controls the trade-off between current and past gradients. Such updates claim to stabilize the training and prevent the parameters from getting stuck at local minima.

To incorporate the idea of Momentum within the KOALA framework, one can simply introduce the state velocities and define the following dynamics
\begin{align}
    x_k & = x_{k - 1} + p_{k - 1} + w_{k-1} \\
    p_k & = \kappa p_{k - 1} + u_{k-1} \\
    \hat L^\text{target}_{k} & = \hat L_{k}(x_k) + v_k,
\end{align}
where $p_k\in\mathbb{R}^n$ and $u_{k-1}$ is a zero-centered Gaussian random variable.

One can rewrite these equations again as Kalman filter equations by combining the parameters $x_k$ and the velocities $p_k$ into one state vector $\bar x_k = [x_k, p_k]$ and similarly for the state noise $\zeta_{k-1} = [w_{k-1}, u_{k-1}]$. This results in the following dynamical system
\begin{align}
    \bar x_k & = F \bar x_{k - 1} + \zeta_{k-1} \\
    \hat L^\text{target}_{k} & = \hat L_{k}(\Pi \bar x_k) + v_k,\label{eq:scalarobs}
\end{align}
where $F = \scriptsize{\begin{bmatrix}
          1 & 1 \\
          0 & \kappa
           \end{bmatrix}} \otimes I_{n \times n}$, $\Pi = \begin{bmatrix}
         1 & 0
    \end{bmatrix} \otimes I_{n \times n} $, and $\otimes$ denotes the Kronecker product.
Similarly to the KOALA-V algorithm, we also aim to drastically reduce the dimensionality of the posterior covariance, which now is a $2n \times 2n$ matrix. We approximate $P_k$ with the following form $\scriptsize{\begin{bmatrix}
        \sigma^2_{x,k} & \sigma^2_{c,k} \\
        \sigma^2_{c,k} & \sigma^2_{p,k}
    \end{bmatrix}}\otimes I_{n \times n} $, where $\sigma^2_{x,k}$, $\sigma^2_{c,k}$, $\sigma^2_{p,k}$ are scalars.
In this formulation we have that
$H_k = \scriptsize{\begin{bmatrix}
         \nabla \hat L_{k}^\top & \boldsymbol{0}^\top
    \end{bmatrix}}$
and thus our approximation for the Kalman update of the posterior covariance will use
\begin{align}
    H_k^\top H_k \approx \left[\begin{array}{cc}
         |\nabla \hat L_{k}|^2 I_{n \times n} & \boldsymbol{0} \\
         \boldsymbol{0} & \boldsymbol{0}
    \end{array}\right].
\end{align}
The remaining equations follow directly from the application of Table~\ref{t:kalmanfilter}.
We call this method the \emph{KOALA-M} (Momentum) algorithm.

\subsection{Estimation of the Measurement and State Noise} 

In the KOALA framework we model the noise in the observations and the state transitions with zero-mean Gaussian variables with covariances $R_k$ and $Q_k$ respectively. So far, we assumed that these covariances were given and constant. However, they can also be estimated online, and lead to more accurate state and posterior covariance estimates. For $R_k$ we use the following running average
\begin{align}
    R_k = \beta_R R_{k - 1} + (1 - \beta_R)\frac{1}{|{\cal C}_k|} \sum_{i \in {\cal C}_k} \left(\ell(\xi_i; x_k)-\hat L^\text{target}_k\right)^2,
\end{align}
where we set $\beta_R = 0.9$.
Similarly, for the covariance $Q_k\doteq\text{diag}\{ q^2_{x,k}I_{n \times n}, q^2_{p}I_{n \times n}\}$, 
the online update for $q_{x,k}^2$ is
\begin{align}
    x_{\text{avg}} & = \beta_x x_{\text{avg}} + (1 - \beta_x) x_{k - 1} \\
    q_{x,k}^2 & = \frac{1}{n} | x_{k - 1} - x_{\text{avg}} | ^2,
\end{align}
where we set $\beta_x = 0.9$.
This adaptivity of the noise helps both to reduce the number of hyper-parameters and to obtain better convergence.

\subsection{Learning Rate Scheduling}

In both the KOALA-V and the KOALA-M algorithms, the update equation for the state estimate needs $\hat L^\text{target}_{k}$ (see \eg, eq.~\eqref{eq:vanillakalman_update}).
Thanks to Theorem~\ref{thm:convergence1} (see next section), we have the option to change $\hat L^\text{target}_{k}$ progressively with the iteration time $k$. For instance, we could set $\hat L^\text{target}_{k} = (1-\varepsilon_k)\hat L_{k}(x_k)$, for some choice of the sequence $\varepsilon_k$. Using this term is equivalent to setting $\hat L_k^\text{target} = 0$ and scaling the learning rate by $\varepsilon_k$ in eq.~\eqref{eq:vanillakalman_update},
as in many SGD implementations \cite{goffin1977lrschedule, loshchilov_ICLR17SGDR}.
Notice the very different interpretation of the schedule in the case of KOALA, where we gradually decrease the target risk.

\subsection{Layer-wise Approximations} 
Let us consider the optimization problem specifically for large neural networks. 
We denote with $B$ the number of layers in a network. Next, we substitute the scalar observation eq.~\eqref{eq:scalarobs} with a $B$-dimensional vector of identical observations. The $i$-th entry  in  this $B$-dimensional observation vector depends  only  on  the variables of the $i$-th block of the network, while the other variables are frozen. Thus, while in the original definition the measurement equation had $H_k$ as an $n$-dimensional vector, under the proposed approximation $H_k$ is a $B \times n$ block-diagonal matrix.
Under these assumptions, the update equation (\ref{eq:vanillakalman_update}) for both the \emph{KOALA-V} and the \emph{KOALA-M} algorithm will split into $B$ layer-wise equations, where each separate equation incorporates only the gradients with respect to the parameters of a specific layer. Additionally to this, now the matrix $H_k^\top H_k S_k^{-1}$ also yields $B$ separate blocks (one per observation), each of which gets approximated by the corresponding largest block eigenvalue. Finally, the maximum of these approximations gives us the approximation of the whole matrix
\begin{align}
    H_k^\top H_k S_k^{-1} \approx 
        \max_{1 \leq i \leq B}{\frac{|\nabla_{b_i} \hat L_{k}|^2}{S_k^{(i)}}},
\end{align}
where $b_i$ is the subset of parameters corresponding to the $i$-th layer and $S_k^{(i)}$ is the innovation covariance corresponding to only the $i$-th measurement. We observe that this procedure induces better convergence in training. For more details, see the supplementary material.

\section{Convergence Analysis}

Our convergence analysis for KOALA builds on the framework introduced in \cite{bertsekas2000gradient}. The analysis is based on a general descent algorithm
\begin{align}
    x_{k + 1} = x_k + \gamma_k (s_k + u_k),
\end{align}
where $\gamma_k$ is the step size, $s_k$ is a descent direction and $u_k$ is a noise term. Here, $s_k$ is related to the gradient of the empirical risk $\hat L$ and $u_k$ satisfies some regularity conditions.
In our algorithm, we also skip all update steps when the norm of the gradient of a minibatch loss is lower than a threshold. This is because such observations provide almost no information to the state. Because of this rule we can thus guarantee that $|\nabla \hat L_{k}(x)|\ge g$ for some $g>0$.
Given these settings, we analyze the evolution of $\hat P_k$, showing that it stays within two positive bounds. Further, we show that the gradient of the loss goes to $0$ as $k\rightarrow \infty$. This result is formalized for the KOALA-V and summarized in the following theorem with two choices for the target risks.
\begin{theorem}
\label{thm:convergence1}
Let $\hat L(x)$ be a continuously differentiable function and $\nabla \hat L(x)$ be Lipschitz-continuous. Assume that $g \le |\nabla \hat L_{k}(x)|\le G$ for all $x$ and $k$, where $g, G > 0$ and $\hat L_k(x)$ is the minibatch loss.
Let us choose the target risk
\begin{eqnarray}
    \text{(a)} \qquad \qquad & \hat L^\text{target}_{k} = &  \hat L_{k}(x_k) - \varepsilon_k, \qquad \qquad \\
    \text{(b)} \qquad \qquad & \hat L^\text{target}_{k} = & (1-\varepsilon_k)\hat L_{k}(x_k), \qquad \qquad
\end{eqnarray}
where $\varepsilon_k$ is any sequence satisfying $\sum_{k=0}^{\infty} \varepsilon_k = \infty$ and $\sum_{k=0}^{\infty} \varepsilon_k^2 < \infty$.
Then $\hat L(x_k)$ converges to a finite value and $\lim \sup_{k \to \infty} |\nabla \hat L(x_k)| \leq g$.
\end{theorem}
\begin{proof}
See supplementary material.
\end{proof}

\section{Ablations}

In this section we ablate the following features and parameters of both \emph{KOALA-V} and \emph{KOALA-M} algorithms: the dynamics of the weights and velocities, the initialization of the posterior covariance matrix and the adaptivity of the state noise estimators. In some ablations we also separately test the \emph{KOALA-M} algorithm with adaptive $Q_k$. Also, we show that our algorithm is relatively insensitive to different batch sizes and weight initializations.

We evaluate our optimization methods by computing the test performance achieved by the model obtained with the estimated parameters. Although such performance may not uniquely correlate to the performance of our method, as it might be affected also by the data, model and regularization, it is a useful indicator.
In all the ablations, we choose the classification task on CIFAR-100~\cite{cifar} with ResNet18~\cite{he2016resnet}. We train all the models for $100$ epochs and decrease the learning rate by a factor of {$0.2$} every $30$ epochs. 

For the last two ablations and in the Experiments section, we use the \emph{KOALA-M} algorithm with $\kappa = 0.9$, adaptive $R_k$ and $Q_k$, initial posterior covariance parameters $\sigma_{x,0}^2 = \sigma_{p,0}^2 = 0.1$ and $\sigma_{c,0}^2 = 0$. 

\noindent\textbf{Impact of the state dynamics and noise adaptivity.} We compare the \emph{KOALA-V} algorithm (\ie, constant  dynamics) to the \emph{KOALA-M} (\ie, with velocities). Additionally, we ablate the $\kappa$, \ie, the decay rate of the velocities. The results are shown in Table~\ref{t:alpha}. We observe that the use of velocities with a calibrated moment has a positive impact on the estimated parameters. Further, with adaptive noise estimations there is no need to set their initial values, which reduces the number of hyper-parameters to tune.

\begin{table}[t]
	\begin{center}
	\caption{Ablation of the  state dynamics and noise adaptivity.}\label{t:alpha}
	\setlength{\tabcolsep}{3pt}
	\begin{tabular}{c@{\hspace{2.5em}}c@{\hspace{2.5em}}c@{\hspace{2.5em}}c}  \toprule
		KOALA   & $\kappa$ &  Top-1 Err. & Top-5 Err. \\ \hline
		V & - & {24.17} & {7.06} \\ \midrule
		M  & 0.50 &  {27.20} & {8.29} \\
	    M  & 0.90 & {23.38} & {6.77} \\ \midrule
		M (adapt. $Q_k$) & 0.50 & 28.25  & 8.91 \\
		M (adapt. $Q_k$) & 0.90 & 23.39 & 6.50\\
		 \bottomrule
	\end{tabular}
	\end{center}
\end{table}

\noindent\textbf{Posterior covariance initialization.} The KOALA framework requires to initialize the matrix $P_0$. In the case of the \emph{KOALA-V} algorithm, we approximate the posterior covariance with a scaled identity matrix, \ie, $P_k = \sigma^2_k I_{n \times n}$, where $\sigma_k \in \mathbb{R}$. In the case of \emph{KOALA-M}, we approximate $P_k$ with a $2 \times 2$ block diagonal matrix {with $\sigma_{x,0}^2 I_{n \times n}$ and $\sigma_{p,0}^2 I_{n \times n}$ on the diagonal, where $\sigma_{x,0}, \sigma_{p,0} \in \mathbb{R}$}. In this section we ablate $\sigma_{x,0}$, $\sigma_{p,0}$ and $\sigma_{0}$ to show that the method quickly adapts to the observations and the initialization of $P_k$ does not have a significant impact on the final accuracy achieved with the estimated parameters. The results are given in Table~\ref{t:sigma}.  


\begin{table}[t]
	\begin{center}
    \caption{Ablation of the posterior covariance initialization.}\label{t:sigma}
	\begin{tabular}{@{}c c c c c@{}} \toprule
		Parm.  & Value & KOALA  &  Top-1 Err. & Top-5 Err. \\ \hline
		\multirow{3}{*}{$\sigma_0^2$} & 0.01 &  V & {24.42} & {7.23} \\
		& 0.10  & V & {24.17} & {7.06} \\ 
		& 1.00 & V & {24.69} & {7.36} \\ \hline	
        
        \multirow{3}{*}{$\sigma_{x,0}^2$} 
		& 0.01 & M (adapt. $Q_k$) & 23.67  & 6.81  \\ 
		& 0.10 & M (adapt. $Q_k$) & 23.39 & 6.50 \\
		& 1.00 & M (adapt. $Q_k$) & 23.82  & 6.53 \\ \hline
		
		\multirow{3}{*}{$\sigma_{p,0}^2$} 
		& 0.01 & M (adapt. $Q_k$) & 23.37 & 7.13 \\
		& 0.10 & M (adapt. $Q_k$) & 23.39 & 6.50 \\
		& 1.00 & M (adapt. $Q_k$) & 24.24 & 7.40 \\ \bottomrule
	\end{tabular}
    \end{center}
\end{table}

\noindent\textbf{Batch size.} Usually one needs to adapt the learning rate to the chosen minibatch size. In this experiment, we change the batch size in the range $[32, 64, 128, 256]$ and show that {KOALA-M} adapts to it naturally. Table~\ref{t:batch_size} shows that the accuracy of the model does not vary significantly with a varying batch size, which is a sign of stability.

\begin{table}[t]
    \center
    \footnotesize
	\caption{Ablation of the batch size used for training. Classification error on CIFAR-100 with ResNet18.}\label{t:batch_size}
	\begin{tabular}{ c c c} \toprule
		  Batch Size  &  Top-1 Error & Top-5 Error \\ \hline
		 32 & 24.59 & 7.13 \\
		 64 & 23.11 & 6.93 \\	
		 128 & 23.39 & 6.50  \\	
		 256 & 24.34 & 7.59 \\ \bottomrule	
	\end{tabular}
\end{table}

\noindent\textbf{Weight initialization.} Like with the batch size, we use different initialization techniques to show that the algorithm is robust to them. We apply the same initializations to SGD for comparison. We test Kaiming Uniform~\cite{Kaiming_2015_Initialization}, Orthogonal~\cite{Saxe_2014_Initialization}, Xavier Normal~\cite{Glorot_2010_Xavier}, Xavier Uniform~\cite{Glorot_2010_Xavier}. The results are shown in Table~\ref{t:weight_init}.
\begin{table}[t]
    \center
    \footnotesize
	\caption{Ablation of different weight initializations. Classification error on CIFAR-100 with ResNet18. }\label{t:weight_init}
	\begin{tabular}{c c c c} \toprule
		Initialization  & Optimizer  &  Top-1 Error & Top-5 Error \\ \hline
		\multirow{2}{*}{Xavier-Normal} & SGD & 26.71 & 7.59 \\
		 & {KOALA-M}  & 23.34 & 6.78 \\ \hline
		\multirow{2}{*}{Xavier-Uniform} & SGD & 26.90 & 7.97 \\
		& {KOALA-M}  & 23.40 &  6.85 \\ \hline
		\multirow{2}{*}{Kaiming-Uniform} & SGD & 27.82 & 7.95 \\
		& {KOALA-M}  & 23.35 & 6.76 \\ \hline
		\multirow{2}{*}{Orthogonal} & SGD & 26.83 & 7.59 \\
		& {KOALA-M}  & 23.27 & 6.63 \\ \bottomrule
	\end{tabular}
\end{table}

\section{Experiments}

We evaluate {KOALA-M} on different tasks, including image classification (on CIFAR-10, CIFAR-100 and ImageNet~\cite{Russakovsky_2015_IJCV}), generative learning and language modeling.
For all these tasks, we report the quality metrics on the validation sets to compare {KOALA-M} to the commonly used optimizers.
We find that {KOALA-M} outperforms or is on par with the existing methods, while requiring fewer hyper-parameters to tune. We will make our code available.

\begin{table}[t]
	\center
	\footnotesize
	\caption{Results on CIFAR-10, CIFAR-100 and ImageNet-32 datasets for 100 and 200 epochs runs.} \label{t:classification}
	\begingroup
    \setlength{\tabcolsep}{1.3pt}
	\begin{tabular}{c c c c c c c} \toprule
		 \multicolumn{3}{c}{} & \multicolumn{2}{ c}{100-epochs} & \multicolumn{2}{c}{200-epochs}   \\
		\midrule
		  & &  & \multicolumn{4}{c}{Error}\\
		Dataset  &Architecture & Method & Top-1 & Top-5 & Top-1   & Top-5  \\ \bottomrule	
		 \multirow{9}{*}{CIFAR-10} & \multirow{3}{*}{ResNet-18} & SGD & \textbf{5.60}  & \textbf{0.16}  &  7.53  & 0.29 \\
		 & & Adam & 6.58 & 0.28 & 6.46  & 0.28\\
		 &  &  {KOALA-M}  & 5.69 & 0.21 & \textbf{5.46} & \textbf{0.25}\\ \cline{2-7} 
		 & \multirow{3}{*}{ResNet-50} & SGD & \textbf{6.37} & \textbf{0.19} & 8.10 & 0.27 \\ 
		 & & Adam& 6.28 & 0.24 & \textbf{5.97} & 0.28 \\
		 & & {KOALA-M} & 7.29 & 0.24 & 6.31 & \textbf{0.13} \\ \cline{2-7}
		  & \multirow{3}{*}{W-ResNet-50-2} & SGD & 6.08 & \textbf{0.15} & 7.60  & 0.24 \\
		 & &Adam & \textbf{6.02} & 0.19 & 5.90  & 0.26 \\
		 & & {KOALA-M} & 6.83 & 0.19 & \textbf{5.36}  & \textbf{0.12}\\ \midrule
		 \multirow{9}{*}{CIFAR-100} & \multirow{3}{*}{ResNet-18} & SGD & 23.50 & 6.48 & 22.44  & \textbf{5.99} \\
		& & Adam & 26.30 & 7.85 & 25.61  & 7.74\\
		&  &  {KOALA-M}  & \textbf{23.38} &\textbf{ 6.70} & \textbf{22.22}  & 6.13\\ \cline{2-7} 
		& \multirow{3}{*}{ResNet-50} & SGD & 25.05 &  6.74 & 22.06  & 5.71 \\ 
		& & Adam& 24.95 & 6.96 & 24.44  & 6.81 \\
		& & {KOALA-M} & \textbf{22.34} & \textbf{5.96} & \textbf{21.03}  & \textbf{5.33} \\ \cline{2-7}
		& \multirow{3}{*}{W-ResNet-50-2} & SGD & 23.83 & 6.35 & 22.47  & 5.96 \\
		& &Adam & 23.73 & 6.64 & 24.04  & 7.06 \\
		& & {KOALA-M} & \textbf{21.25} & \textbf{5.35} & \textbf{20.73}  & \textbf{5.08}\\ \midrule
		 \multirow{2}{*}{ImageNet-32} & \multirow{2}{*}{ResNet-50} & SGD & \textbf{34.07} & \textbf{13.38} & -  & - \\
		& & {KOALA-M} & 34.99 & 14.06 & -  & -\\ 		\bottomrule	 
	\end{tabular}
	\endgroup
\end{table}

\noindent\textbf{CIFAR-10/100 Classification.} We first evaluate {KOALA-M} on CIFAR-10 and CIFAR-100 using the popular ResNets~\cite{he2016resnet} and WideResNets~\cite{zagoruyko2016wideresnets} for training. 
We compare our results with the ones obtained with commonly used existing optimization algorithms, such as SGD with Momentum and Adam. 
For SGD we set the momentum rate to $0.9$, which is the default for many popular networks, and for Adam we use the default parameters $\beta_1=0.9, \beta_2=0.999, \epsilon=10^{-8}$. 
In all experiments on CIFAR-10/100, we use a batch size of $128$ and basic data augmentation (random horizontal flipping and random cropping with padding by $4$ pixels). 
For each configuration we have two runs for $100$ and $200$ epochs respectively. 
For SGD we start with a learning rate equal to $0.1$, for Adam to $0.0003$ and $1.0$ for {KOALA-M}. 
{
For the $100$-epochs run on CIFAR-10 (CIFAR-100) we decrease the learning rate by a factor of $0.1$ $(0.2)$ every $30$ epochs. 
For $200$-epochs on CIFAR-10 we decrease the learning rate only once at epoch $150$ by the factor of $0.1$. 
For the $200$-epoch training on CIFAR-100 the learning rate is decreased by a factor of $0.2$ at epochs $60$, $120$ and $160$.
}
For all the algorithms, we additionally use a weight decay of $0.0005$.
To show the benefit of using {KOALA-M} for training on classification tasks, we report the Top-1 and Top-5 errors on the validation set.
For both the $100$-epochs and $200$-epochs configurations, we report the mean error among $3$ runs with $3$ different random seeds. {Note that the $100$/$200$-epochs configurations are not directly comparable due to the different learning rate schedules.} 
The results are reported in Table~\ref{t:classification}.
For more comparisons and training plots see the Supplementary material.
 
\noindent\textbf{ImageNet Classification.} Following \cite{ADAM_W}, we train a ResNet50 \cite{he2016resnet} on $32\times 32$ downscaled images with the most common settings: $100$ epochs of training with learning rate decrease of $0.1$ after every $30$ epochs and a weight decay of $0.0001$. We use random cropping and random horizontal flipping during training and we report the validation accuracy on single center crop images. 
As shown in Table~\ref{t:classification}, our model achieves a comparable accuracy to SGD, but without any task-specific hyper-parameter tuning. 

\noindent\textbf{Comparison to more recent algorithms.} We compare {KOALA-M} with a wider range of optimizers on the CIFAR-100 classification with ResNet50 in the $100$-epochs configuration. We used the same learning rate schedule as in the previous section and set the hyperparameters for the other algorithms to the ones reported by the authors. For Yogi~\cite{yogi} we set the learning rate to $10^{-2}, \beta_1 = 0.9, \beta_2 = 0.999$ and $\epsilon = 10^{-3}$, as suggested in the paper. For Adamax~\cite{ICLR_KingmaB_2014}, AdamW~\cite{ADAM_W}, AdamP~\cite{ADAMP_SGDP} and Amsgrad~\cite{ADAM_NC} we use the same hyperparameters as for Adam. For Fromage~\cite{NEURIPS2020_f4b31bee} we set the learning rate to $10^{-2}$, as suggested on the project's github page\footnote{\url{https://github.com/jxbz/fromage#voulez-vous-du-fromage}}. For Adabelief~\cite{NEURIPS2020_Zhuang} we follow the hyperparameters reported in the official implementation\footnote{\url{https://github.com/juntang-zhuang/Adabelief-Optimizer#hyper-parameters-in-pytorch}} . The results are shown in Table~\ref{t:many_optimizers}.

\begin{table}[t]
    \center
    \footnotesize
    \caption{Comparison of different optimizers on CIFAR-100 classification task with 100-epochs configuration. Mean errors across 3 runs with different random seeds are reported.}\label{t:many_optimizers}
    \setlength{\tabcolsep}{1.3pt}
	\begin{tabular}{l c c} \toprule
	    Optimizer  &  Top-1 Err. & Top-5 Err. \\ \hline
	    Yogi~\cite{yogi} & 33.99 & 10.90 \\
	    Adamax~\cite{ICLR_KingmaB_2014} & 32.42 & 10.74 \\
	    AdamW~\cite{ADAM_W} & 27.23 & 7.98 \\
	    AdamP~\cite{ADAMP_SGDP} & 26.62 & 7.61 \\
	    Amsgrad~\cite{ADAM_NC} & 25.27 & 6.78 \\
	    Fromage~\cite{NEURIPS2020_f4b31bee} & 24.65 & 6.71 \\
	    Adabelief~\cite{NEURIPS2020_Zhuang} & 23.07 & 6.05 \\
	    {KOALA-M} & \textbf{22.34} & \textbf{5.96} \\
		  \bottomrule	
	\end{tabular}
\end{table}

In Table~\ref{t:exp}, we show that {KOALA-M} is compatible with such auxiliary methods as Lookahead (LA)~\cite{zhang2019lookahead} and SWA~\cite{izmailov2018averaging}. For LA we used SGD and Adam with initial learning rates equal to 0.1 and 0.0003 respectively as the inner optimizers and set the hyperparameters $\alpha$ and $k$ to 0.8 and 5 respectively, as suggested by the authors. We used SWA with both SGD and Adam inner optimizers averaging every 5 epochs starting from epoch 75. Additionally, we apply LA and SWA to {KOALA-M}. All experiments are for CIFAR-100 classification with ResNet50. Training is done in 100-epochs configuration.

{\renewcommand{\arraystretch}{1.2}
\begin{table}[t]
    \center
    \caption{Classification error on CIFAR-100 with ResNet50. Average of 3 runs with different random seeds is reported.}\label{t:exp}
	\begin{tabular}{
		@{\hspace{.5em}} c@{\hspace{.5em}} |@{\hspace{.5em}} c@{\hspace{.5em}} c@{\hspace{.5em}} c@{\hspace{.5em}} c@{\hspace{.5em}} c@{\hspace{.5em}} |
		@{\hspace{.5em}}c@{\hspace{.5em}} c@{\hspace{.5em}}}
	    \hline
        \begin{turn}{90}Optimizer\end{turn}  &
		\begin{turn}{90}LA(SGD)\end{turn}  &
		\begin{turn}{90}LA(Adam)\end{turn}  &
		\begin{turn}{90}SWA(SGD)\end{turn}  & 
		\begin{turn}{90}SWA(Adam)~\end{turn}  & 
		\begin{turn}{90}{KOALA-M}\end{turn}  &
		\begin{turn}{90}SWA({KOALA-M})\end{turn}  &
		\begin{turn}{90}LA({KOALA-M})\end{turn}  \\
		\hline
        Top-1 &
        23.36 &
        24.90 &
        24.24 &
        24.49 &
        \textbf{22.34} &
        22.29 &
        \textbf{21.70} \\
        \hline
	\end{tabular}
\end{table}
}

\noindent\textbf{Memory and time complexity.} {KOALA-M} needs at most $2\times$ the size of the network in additional memory for storing $x_{\text{avg}}$ and state velocities $p_k$.
Also, since we do not store the full state covariance matrices and use at most $2 \times 2$ matrices in update equations, the computational complexity of our algorithm is \textbf{linear} with respect to the network parameters. For numerical results see the Supplementary material.

\noindent\textbf{GANs and language modeling.} {KOALA-M} also works well for training GANs~\cite{goodfellow2014generative} and on NLP tasks. For numerical results, see the supplementary material.

\section{Conclusions}

We have introduced KOALA, a novel Kalman filtering-based approach to stochastic optimization. KOALA is suitable to train modern neural network models on current large scale datasets with high-dimensional data. The method can self-tune and is quite robust to wide range of training settings. Moreover, we design KOALA so that it can incorporate optimization dynamics such as those in Momentum and Adam, and learning rate schedules. The efficacy of this method is demonstrated on several experiments in image classification, image generation and language processing.


{
\fontsize{9.8pt}{10.8pt} \selectfont
\bibliography{references}
}

\end{document}